\documentclass[runningheads]{llncs}
\usepackage[T1]{fontenc}
\usepackage{tabularx}
\usepackage{booktabs}
\usepackage{graphicx}
\usepackage{caption}
\usepackage{amsmath}
\usepackage{hyperref}
\usepackage{graphicx} % for including graphics
\usepackage{lipsum}  % for generating dummy text
\usepackage{wrapfig}
\usepackage{enumitem}
\usepackage{changepage}

\usepackage[backend=biber,style=numeric,sorting=none]{biblatex} %Import biblatex package 
\newcommand{\changefont}{\fontsize{7pt}{10pt}\selectfont}

\begin{document}
\title{Generative Adversarial Networks for Imputing Sparse Learning Performance \thanks{This paper was officially accepted on June 24 by the International Conference on Pattern Recognition (ICPR) 2024.}} 
\author{Liang Zhang \inst{1,2} \and
Mohammed Yeasin \inst{1,2} \and
Jionghao Lin \inst{3, 4} \and Felix Havugimana \inst{2} \and Xiangen Hu \inst{1,2,5}} 
\authorrunning{L. Zhang et al.}
\institute{Institute for Intelligent Systems, University of Memphis, Memphis, TN 38152, USA \and Department of Electrical and Computer Engineering, University of Memphis, Memphis, TN 38152, USA \\
\email{\{lzhang13,myeasin,fhvgmana\}@memphis.edu}
\and
Human-Computer Interaction Institute, Carnegie Mellon University, Pittsburgh, PA 15213, USA 
\and
Centre for Learning Analytics, Monash University, Melbourne, VIC 3800, Australia \\
\email{jionghal@cs.cmu.edu} \and
Department of Applied Social Sciences, Hong Kong Polytechnic University, Hong Kong, PR China \\
\email{xiangen.hu@polyu.edu.hk}}
\maketitle              % typeset the header of the contribution
\begin{abstract} Learning performance data, such as correct or incorrect responses to questions in Intelligent Tutoring Systems (ITSs) is crucial for tracking and assessing the learners' progress and mastery of knowledge. However, the issue of data sparsity, characterized by unexplored questions and missing attempts, hampers accurate assessment and the provision of tailored, personalized instruction within ITSs. This paper proposes using the Generative Adversarial Imputation Networks (GAIN) framework to impute sparse learning performance data, reconstructed into a three-dimensional (3D) tensor representation across the dimensions of learners, questions and attempts. Our customized GAIN-based method computational process imputes sparse data in a 3D tensor space, significantly enhanced by convolutional neural networks for its input and output layers. This adaptation also includes the use of a least squares loss function for optimization and aligns the shapes of the input and output with the dimensions of the questions-attempts matrices along the learners' dimension. Through extensive experiments on six datasets from various ITSs, including AutoTutor, ASSISTments and MATHia, we demonstrate that the GAIN approach generally outperforms existing methods such as tensor factorization and other generative adversarial network (GAN) based approaches in terms of imputation accuracy. This finding enhances comprehensive learning data modeling and analytics in AI-based education. 

\keywords{Learning Performance Data \and Data Imputation \and Generative Adversarial Imputation Networks \and Generative Artificial Intelligence Model \and Intelligent Tutoring System} 
\end{abstract}
\section{Introduction}

The learning performance data, recorded during ITS interactions, documents the sequence of questions-answering activities, tracking the identifications of questions and cataloging the attempts and performance responses made by different learners. However, real-world learning performance data is often incomplete and sparse, with unexplored questions and limited attempts posing challenges for data analysis and modeling. Multiple reasons contribute to data sparsity, including participant's dropout from learning tasks \cite{psathas2023predictive}, learner disengagement due to off-task behavior \cite{baker2007modeling}, random data loss from design and operational errors \cite{saarela2017automatic}, biases within sample groups \cite{greer2016evaluation}, among others. This sparsity hinders a comprehensive understanding and assessment of learning performance. Such limitations hinder AI-powered educational systems from effectively delivering educational content, especially in their capabilities to track learning processes, monitor advancement, and gauge learners' mastery of knowledge through their performance data. Thus, accurately imputing sparse learning performance data is critical for advancing learning analytics and modeling, which facilitates the comprehensive exploration of learning insights and ultimately enhancing learner progress within ITSs. 

Although traditional data imputation methods (e.g., indicator or mean imputation \cite{batista2003analysis,donders2006gentle}, regression imputation \cite{zhang2016missing}, and multiple imputation \cite{rubin1978multiple}) have proven effective in the literature, they provide a cost-effective solution that avoids labor-intensive experiments and leverages observed data to estimate unobserved data, capitalizing on underlying patterns and characteristics \cite{rubin1977assignment}. Indicator or mean imputation may introduce bias by oversimplifying missing data complexities \cite{batista2003analysis,donders2006gentle}. Regression imputation often fails to capture the full spectrum of the underlying data structure \cite{zhang2016missing}. Multiple imputation may not adequately address complex, high-dimensional correlations \cite{seaman2012multiple}. Recently, generative AI models, specifically Generative Adversarial Networks (GANs) \cite{goodfellow2014generative}, have demonstrated remarkable success in handling data sparsity through the reconstruction mechanism \cite{yoon2018gain,dong2021generative}, achieving higher accuracy and effectively addressing those issues in traditional data imputation methods. A notable GAN-based model, Generative Adversarial Imputation Nets (GAIN) demonstrated its effectiveness on imputing human health data \cite{yoon2018gain}. The GAIN model extends GAN structure by conditioning the generator on observed data and using a hint mechanism to enhance the discriminator's accuracy in identifying missing data patterns \cite{yoon2018gain,zhang2023systematic}. Further research has shown GAIN's superior performance in diverse datasets and applications, from healthcare to machine health monitoring, validating its effectiveness over traditional methods like MICE and missForest \cite{yoon2018gain,dong2021generative,hu2022fault}.

Despite its impressive imputation performance in prior studies, GAIN's potential for imputing missing data in sparse learning performance datasets within ITSs remains unexplored.  The complex nature of learning performance data, characterized by individual learners, questions, and attempts, presents significant challenges for generative models in data imputation. These challenges include achieving higher accuracy based on existing data distribution and handling the complexities of data interactions and variations across different attempts. Therefore, how can we effectively represent learning performance data to ensure compatibility with the GAIN framework? Additionally, what modifications are necessary to facilitate accurate predictions through specialized computations and algorithms tailored for learning performance scenarios, considering the stability of the models amid dynamic changes in learning performance data? 

In response to these challenges in learning performance data, our study aims to perform the data imputation for the sparse learning performance data using the GAIN framework, enriched with detailed revisions. We are guided by the following two \textbf{R}esearch \textbf{Q}uestions: 
\begin{itemize}
    \item \textbf{RQ 1}: How effectively does the GAIN-based method impute sparse learning performance data in ITSs compared to established baseline methods?
    \item \textbf{RQ 2}: How does the stability of the GAIN-based model's performance vary with changes in the number of attempts influencing the sparsity levels of learning performance data? 
\end{itemize} 

The generative AI model, GAIN, was leveraged to impute the sparse learning performance data, with an additional focus on exploring the stability of GAIN. Therefore, this study's contributions are twofold:
\begin{itemize} 
    \item[$\bullet$] It enhances the accuracy of imputing learning performance data, thereby enriching data representation for more detailed analytics and modeling.
    \item[$\bullet$] The findings are expected to provide valuable imputation methods for comprehensively tracing and assessing learners' progress within AI-based educational systems like ITS. 
\end{itemize} 

\section{Related Work}
\subsection{Addressing Data Sparsity in ITSs}
In AI-based education, many studies have focused on tackling data sparsity in sparse learning performance in ITSs. Chen et al. \cite{chen2018prerequisite} employed the prerequisite concept map for knowledge tracing to mitigate data sparsity. Pandey et al. \cite{pandey2019self} developed the self-attentive mechanism to predict the learner's performance on unanswered questions by analyzing the relevance previously answered questions. Wang et al. \cite{wang2019deep} integrated question-knowledge hierarchies into a deep learning framework to improve predictions despite data sparsity. Despite these advances, challenges persist: (1) high demands for expert effort in mapping and annotating knowledge concepts \cite{novak2006theory}, (2) ignorance of temporal learning dynamics \cite{thai2012factorization}, and (3) disruption of sequential learning effects even with some methodological recognition \cite{conway2001sequential,conway2012}. 

In addressing data sparsity in ITSs, tensor factorization has also played a pivotal role by leveraging multidimensional relationships to enhance prediction accuracy and knowledge representation. This approach has evolved from simple matrix factorization to sophisticated multi-dimensional frameworks that incorporate temporal effects and sequential learning dynamics, significantly improving the understanding and prediction of learner performance \cite{thai2011matrix, thai2012factorization, sahebi2016tensor}. Such advancements in tensor factorization have laid the groundwork for employing more advanced generative models like GAIN, which further enhance data imputation by maintaining the natural multidimensional structure of learning data. This alignment with deep generative models has directly influenced our adoption of GAIN to effectively address the complex challenges of data sparsity in ITSs \cite{yoon2018gain}. 

\subsection{Generative AI Models for Educational Data Imputation}
There have been tremendous progress in generative AI model for educational data imputations in ITSs. Morales-Alvarez et al. \cite{morales2022simultaneous} explored the application of generative models, incorporating structured latent spaces and graph neural network-based architectures, to achieve competitive or superior performance in data imputation for real-world mathematics datasets of Eedi (a leading educational platform which millions of students interact with daily around the globe on diagnostic multiple-choice mathematics questions \cite{boyle2021eedi}), surpassing traditional baselines such as MICE and missForest. Ma et al. \cite{ma2021identifiable} also employed deep generative models to effectively impute data for multiple-choice question learning data, addressing over \(70\%\) missing rates in the Eedi educational dataset for mathematics questions. Zhang et al. \cite{zhang20243dg} investigated the use of generative AI models, GAN and GPT, for data augmentation to address sparse learning performance patterns in adult reading comprehension, finding GAN to provide more stable augmentation across various sample sizes. 

Learning performance data from ITSs such as AutoTutor CSAL, which focuses on reading comprehension \cite{graesser2016reading}, along with ASSISTments \cite{heffernan2014assistments} and MATHia \cite{ritter2007cognitive}, which both focus on math learning for middle and high-school students, document learners' responses as correct or incorrect to a sequence of questions posed by the system. Inspired by GANs' capability to impute missing regions in images by training on vast amounts of image data and filling in missing areas to maintain coherence with the existing context \cite{pathak2016context}, we are motivated to apply a similar data imputation approach to tensor-based learning performance data. There is an ongoing need for effective imputation of learning performance data in ITSs like AutoTutor CSAL, ASSISTments and MATHia to enhance educational technology's ability to track and assess learners' performance comprehensively. 

\section{Methods}
\subsection{Dataset}

In this study, we utilized datasets from three primary sources: AutoTutor CSAL lessons\footnotemark[1]\footnotetext[1]{AutoTutor Moodel Website: \href{https://sites.autotutor.org/}{https://sites.autotutor.org/}; Adult Literacy and Adult Education Website: \href{https://adulted.autotutor.org/}{https://adulted.autotutor.org/}}, ASSISTments\footnotemark[2]\footnotetext[2]{ASSISTments Website: \href{https://new.assistments.org/}{https://new.assistments.org/}} and the MATHia\footnotemark[3]\footnotetext[3]{MATHia Website: \href{https://www.carnegielearning.com/solutions/math/mathia/}{https://www.carnegielearning.com/solutions/math/mathia/}} dataset from mathematics class. The AutoTutor CSAL lessons cover topics such as ``\textit{Cause and Effect}'' (CSAL Lesson 1) and ``\textit{Problems and Solution}'' (CSAL Lesson 2), each comprising 8 to 11 multiple-choice questions designed to test adults' reading comprehension skills. This study was granted ethical approval with the Institutional Review Board (IRB) number: H15257. The ASSISTments dataset cover the lesson topics including ``\textit{Algebra Symbolization Studies}'' (ASSISTments Lesson 1)\footnotemark[4]\footnotetext[4]{Assistments 2008-2009: \href{https://pslcdatashop.web.cmu.edu/DatasetInfo?datasetId=388}{https://pslcdatashop.web.cmu.edu/DatasetInfo?datasetId=388}} and ``\textit{Skill Builder}'' (ASSISTments Lesson 2)\footnotemark[5]\footnotetext[5]{Assistments 2012-2013: \href{https://sites.google.com/site/assistmentsdata/datasets/2012-13-school-data-with-affect?authuser=0}{https://sites.google.com/site/assistmentsdata/datasets/2012-13-school-data-with-affect?authuser=0}}. The MATHia lesson dataset\footnotemark[6]\footnotetext[6]{MATHia 2019-2020: \href{https://pslcdatashop.web.cmu.edu/Project?id=720}{https://pslcdatashop.web.cmu.edu/Project?id=720}} include algebra lessons on ``\textit{Scale Drawings}'' (MATHia Lesson 1) and ``\textit{Analyzing Models of Two-Step Linear Relationships}'' (MATHia Lesson 2). Details such as the total number of learners, the total number of questions, and the maximum number of attempts are summarized in Table~\ref{tab:all_dataset}. 
\begin{table*}[ht!] 
\centering
\renewcommand{\arraystretch}{1}  % Provide more space between table rows, if you prefer
\caption{Dataset for the CSAL AutoTutor, ASSISTments and MATHia lessons} 
\label{tab:all_dataset}
\changefont
\begin{tabularx}{\textwidth}{>{\raggedright\arraybackslash}X >{\raggedright\arraybackslash\hsize=1.45\hsize}X >{\centering\arraybackslash\hsize=.5\hsize}X >{\centering\arraybackslash\hsize=.5\hsize}X >{\centering\arraybackslash\hsize=.5\hsize}X}
\toprule
Dataset & Lesson Topics & \#Learners & \#Questions & \#Attempts \\ \midrule 
CSAL Lesson 1 & Cause and Effect & 118 & 9 & 9 \\
CSAL Lesson 2 & Problems and Solution & 140 & 11 & 5 \\
ASSISTments Lesson 1 & Algebra Symbolization Studies & 318 & 64 & 4 \\
ASSISTments Lesson 2 & Skill Builder & 392 & 20 & 4
\\
MATHia Lesson 1 & Scale Drawings & 500 & 28 & 4 \\
MATHia Lesson 2 & Analyzing Models of Two-Step Linear Relationships & 500 & 6 & 4 \\
\bottomrule 
\end{tabularx}
% \hline\end{tabular}
\end{table*}

\subsection{The 3-D Tensor Representation of Sparse Learning Performance}

We define the 3-D tensor \(\boldsymbol{\mathcal{T}}_{sparse}\) to encapsulate the learning performance records of \(U\) learners on \(N\) questions over a sequence of up to \(M\) attempts, with \(\boldsymbol{\mathcal{T}}_{sparse}\in [0,1,NaN]^{U\times N \times M }\). Here, \(U=max(1,2,3,\cdots,u)\) is the maximum number of learners, \(N=max(1,2,3,\cdots,n)\) the maximum number of questions, and \(M=max(1,2,3,\cdots,m)\) the maximum number of attempts. Each element \(\tau_{uij}\) within \(\boldsymbol{\mathcal{T}}_{sparse}\) encodes the learning performance as 1 for correct answer, 0 for an incorrect answer, and \(NaN\) to signify unobserved data for a specific question's performance at certain attempt. 

\subsection{The Proposed GAIN-based Imputation Architecture}
\begin{figure}
\includegraphics[width=1\textwidth]{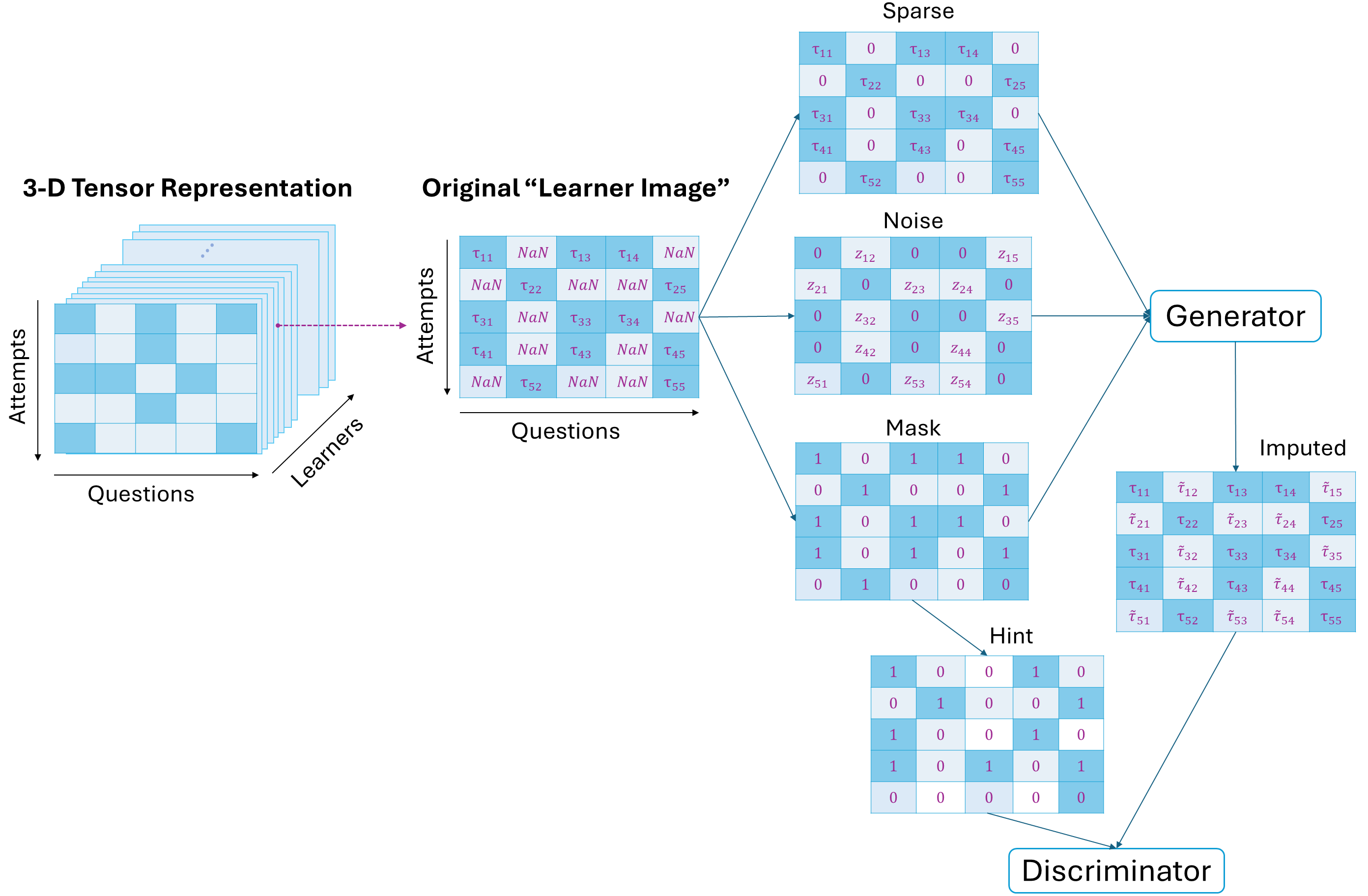}
\centering
\caption{The proposed GAIN-based imputation architecture for sparse learning performance  \cite{yoon2018gain}.} \label{fig:gain}
\end{figure}

Consider the \(\boldsymbol{\mathcal{T}}_{sparse}\), representing the learning performance of all learners. This tensor comprises layers along the learner dimension, represented as \(\boldsymbol{\mathcal{T}}_{sparse}=(\mathcal{T}_{l_{1}},\mathcal{T}_{l_{2}}, \cdots, \mathcal{T}_{l_{n}})\). Each layer, akin to a single-channel ``learner image'', is a matrix that encapsulates performance values across different questions and attempts for an individual learner. This is visualized in Fig.~\ref{fig:gain}. 

For each matrix-based layer \(\mathcal{T}_{l} \in (\mathcal{T}_{l_{1}},\mathcal{T}_{l_{2}}, \cdots, \mathcal{T}_{l_{n}})\), each entry \(\tau_{lij}\) in the \(N \times M\) matrix may include the performance values of 0, 1 or \(NaN\) to present the observed data and unobserved data, respectively. One mask matrix \(\mathcal{T}_{l_{mask}}\) is supposed to map the observed and unobserved entries within the matrix \(\mathcal{T}_{l}\), with 1 signifying observed data, and 0 indicates unobserved data. One noise matrix \(\mathcal{Z}\) with dimensions matching \(\mathcal{T}_{l}\), is initialized. These matrices collectively function as inputs to the generator in the GAIN architecture, producing the output \(\mathcal{T}_{lG}=G(\mathcal{T}_{l},\mathcal{T}_{l_{mask}}, (1-\mathcal{T}_{l_{mask}}) \odot \mathcal{Z})\) \cite{yoon2018gain}. Here, the \(\odot\) denotes as Hadamard product, indicating element-wise multiplication. The imputed matrix \(\mathcal{T}_{l_{imputed}}= \mathcal{T}_{l_{mask}} \odot \mathcal{T}_{l} + (1-\mathcal{T}_{l_{mask}}) \odot \mathcal{T}_{lG}\), effectively merges observed and generated data to fill in unobserved entries. Particularly, a hint matrix \(\mathcal{T}_{l_{hint}}\), also matching the dimensions of \(\mathcal{T}_{l}\) and derived from the mask matrix \(\mathcal{T}_{l_{mask}}\), is introduced. It employs a hint rate to specify the conditional probability that a specific entry in \(\mathcal{T}_{l_{imputed}}\) can be observed, given both \(\mathcal{T}_{l_{imputed}}\) and \(\mathcal{T}_{l_{hint}}\). Thereby, the discriminator within the GAIN architecture, formulated as \(D(\mathcal{T}_{l_{imputed}}, \mathcal{T}_{l_{hint}})\), evaluate this as probability \cite{yoon2018gain}. We train \(D(\cdot)\) to maximize the probability of correctly predicting the \(\mathcal{T}_{l_{mask}}\), while the \(G(\cdot)\) is trained to minimize the likelihood of \(D(\cdot)\) correctly predicting \(\mathcal{T}_{l_{mask}}\). So, we introduce the objective function \(V(D,G)\) \cite{yoon2018gain}:
\begin{equation}\label{eq:vali}
\scalebox{0.8}{%
$V(D,G)= E[\mathcal{T}_{l_{mask}}^{T} logD(\mathcal{T}_{l_{imputed}}, \mathcal{T}_{l_{hint}}) + (1-\mathcal{T}_{l_{mask}})^{T} log(1-D(\mathcal{T}_{l_{imputed}}, \mathcal{T}_{l_{hint}}))]$}
\end{equation}

Our proposed imputation architecture incorporates several novel modifications and configurations from the initial GAIN architecture \cite{yoon2018gain}. See below for further details: 
\begin{itemize}
    \item The convolutional layers are employed for both the generator and discriminator, diverging from the original architecture' reliance on dense layers. Five convolutional neural network (CNN) layers \cite{lecun1989handwritten}, excluding the input and output layers, with the ReLU activation function are applied to the output of each layer. 
    \item During the iterative training phase, the observed data from \(\mathcal{T}_{l}\) and the corresponding imputed data from \(\mathcal{T}_{lG}\) are utilized for optimization via the least square loss function, specifically the Root Mean Square Error (RMSE). This method is chosen to not only ensure enhanced stability and superior quality of the generated data \cite{mao2017least,yoon2020gamin} but also align with probability-based predictions of learning performance in peer research on ITSs \cite{yudelson2013individualized,gervet2020deep,pavlik2021logistic}. 
    \item By incorporating a reshape function in the generator's output layer, the shape of generated data \(\mathcal{T}_{lG}\) is flexible adjustment to fit the given ``learner image'' shape, thus accommodating variations across different lesson scenarios without being constrained to a fixed shape, as commonly seen in image-oriented research \cite{goodfellow2014generative,chen2016infogan,wang2021pc}. 
\end{itemize}

The theoretical foundation for understanding the inference logic and model assumptions in our study includes: 
\begin{itemize}
    \item \textbf{Inference Logic.} The entry set within \(\boldsymbol{\mathcal{T}}_{sparse}\) can be categorized into two subsets: \(\mathcal{T}_{observed}\) for existing values (0 and 1) and \(\mathcal{T}_{unobserved}\) for missing ones (\(NaN\)). The inference model, formulated as \(f_{impute}(\mathcal{T}_{unobserved}|\mathcal{T}_{observed})\), is principle for data imputation,  leveraging observed data patterns to impute missing values and predict outcomes \cite{rubin1976inference}.
    \item \textbf{Model Assumptions.} Our imputation model operates under several key assumptions within a tensor-based framework: \textbf{(a)} \textit{Probability-based prediction}: Assumes predicted learning performance is a continuous probability between 0 and 1, indicative of knowledge mastery \cite{baker2008more}. \textbf{(b)} \textit{Latent domain knowledge relations}: Posits that unobservable latent relationships within the domain knowledge implicitly influence knowledge mastery \cite{corbett1994knowledge,essa2016possible}. \textbf{(c)} \textit{Similarity in learning for individual learners}: Suggests a shared relevance and usefulness of knowledge among learners, aiding in predicting knowledge mastery \cite{thai2011factorization,thai2011matrix}. \textbf{(d)} \textit{Performance interactions influenced by sequence effects}: Acknowledges that learners' interactions with sequential questions are shaped by priming and recency effects, affecting comprehension and performance \cite{conway2001sequential,ramscar2016learning}. \textbf{(e)} \textit{Maximum attempt assumption}: Defines a theoretical maximum number of attempts a learner might need, emphasizing the importance of evaluating comprehensive learning states through repeated trials \cite{corbett1994knowledge}. 
\end{itemize} 

\subsection{Baselines}
This study will compare the proposed GAIN-based imputation method against a range of baseline methods, including those from the tensor factorization series and GAN series. Detailed descriptions of these baselines are provided below. 

\textbf{Tensor Factorization:} The basic tensor factorization factorizes the sparse tensor \(\boldsymbol{\mathcal{T}}_{sparse}\) into two components: a learner latent matrix capturing abilities and learning-related features, and a latent tensor representing knowledge during question attempts \cite{zhang2023exploring,zhang20243dg}. A rank-based constraint is used to maintain a generally positive learning trend and accommodate forgetting or slipping \cite{doan2019rank}. This refined method enhances data imputation within tensor-based structures, providing a robust solution for handling sparse data. 

\textbf{CANDECOMP/PARAFAC Decomposition (CPD):} Drawing on the principle of classic CPD \cite{carroll1970analysis,harshman1970foundations}, the sparse tensor \(\boldsymbol{\mathcal{T}}_{sparse}\) is decomposed into three factor tensors that capture learner, attempt and question-related factors in a multidimensional tensor form. A rank-based constraint is additionally applied to enhance the decomposition's accuracy. 

\textbf{Bayesian Probabilistic Tensor Factorization (BPTF):} The BPTF \cite{xiong2010temporal} is employed to approximate the sparse tensor \(\boldsymbol{\mathcal{T}}_{imputed}\) through the decomposition into a sum of outer products of three lower-dimensional factor tensors. This approach leverages Bayesian inference for sampling both the factor tensors and the precision of observed entries, effectively enhancing the model’s capacity to manage data sparsity and uncertainty \cite{xiong2010temporal,morise2019bayesian}. 

\textbf{Generative Adversarial Network (GAN):} At one core of the GAN, the ``learner image'' extracted from \(\boldsymbol{\mathcal{T}}_{sparse}\) (depicted in Fig.~\ref{fig:gain}), constitutes the base input for the GAN. The GAN architecture includes a generator that simulates data resembling observed entries and a discriminator that assesses the authenticity of this generated data \cite{goodfellow2014generative}. It uses a consistent CNN layer configuration and least squares loss for optimization. 

\textbf{Information Maximizing Generative Adversarial Nets (InfoGAN):} The InfoGAN \cite{chen2016infogan} enhances the traditional GAN framework by integrating the noise with two structured latent variables, allowing for the capture of salient, structured semantic features, such as those relating to learner attributes in ITSs (e.g., initial learning ability and learning rate). The generator generates imputed \(\boldsymbol{\mathcal{T}}_{imputed}\) and decodes latent variables. An auxiliary distribution improves the estimation of these variables' posterior, boosting mutual information between latent codes and observations and ensuring that the generated outcomes are meaningfully informed. 

\textbf{AmbientGAN:} AmbientGAN \cite{bora2018ambientgan} is used to impute sparse learning performance data by training on partially observed or corrupted data within a GAN framework. It incorporates a dynamically adjusted Gaussian blur in the measurement process, enabling the discriminator to effectively distinguish between real and generated data measurements and accurately infer the original dataset's true distribution. 

\subsection{Experimental Setup and Evaluation}
The experimental setup for imputing sparse learning performance data incorporates several tailored configurations to optimize model training and evaluation. \textbf{(a)} \textit{Cross-Validation}: A systematic five-cycle, five-fold cross-validation strategy is rigorously employed for each model to ensure consistency and reliability of results. \textbf{(b)} \textit{Varying Attempt Setting:} The stability of models' data imputation performance is tested under various maximum attempt settings to handle different degrees of data sparsity. \textbf{(c)} \textit{Maximum Iterations:} All models are allowed up to 100 iterations to ensure thorough adaptation and learning. \textbf{(d)} \textit{Learning Rate:} A learning rate of either 0.0001 or 0.00001 is selected to promote steady progress and convergence during model training. \textbf{(e)} \textit{Regularization Techniques:} Dropout and Batch Normalization are integrated into the training process of GAN-based methods to prevent overfitting. \textbf{(f)} \textit{Imputation Accuracy Evaluation Metric:} The Root Mean Square Error (RMSE), as referenced in peer papers \cite{xiong2010temporal,sahebi2016tensor,yoon2018gain}, is used to evaluate models’ performance in data imputation. \textbf{(g)} \textit{Measuring Sparsity Level:} The sparsity level of a tensor-based distribution for
learning performance data is computed as the percentage of missing
values in the total number of elements in the distribution. \textbf{(h)} \textit{Correlation Evaluation:} The Spearman correlation coefficient \cite{spearman1961proof} is used to assess the relationship between RMSE and the varying maximum number of attempts. 

\section{Results and Discussion} 
\noindent % Ensures no indentation
\begin{minipage}[t]{0.5\textwidth}
\vspace{0pt} % Aligns the top of the minipage with the top of the right side
\textbf{Data Sparsity Levels.} Fig.~\ref{fig:sparsity} displays the variation in sparsity levels within learning performance data across six lessons, categorized by the maximum number of attempts. The sparsity level for each lesson increases with the number of attempts, suggesting a progressive introduction of missing data or non-responses for learners in learning process. This trend is consistent across all courses, albeit with varying rates of increase. Notably, ``ASSISTments Lesson 2'' exhibits a gradual ascent, recording the highest sparsity levels across all attempts when compared to other lessons. In contrast, ``MATHia Lesson 1'' and ``MATHia Lesson 2'' demonstrates a low- 
\end{minipage}%
\hspace{0cm} % Adds horizontal space between the minipages
\begin{minipage}[t]{0.5\textwidth}
\vspace{1pt} % Aligns the top
\includegraphics[width=1\textwidth]{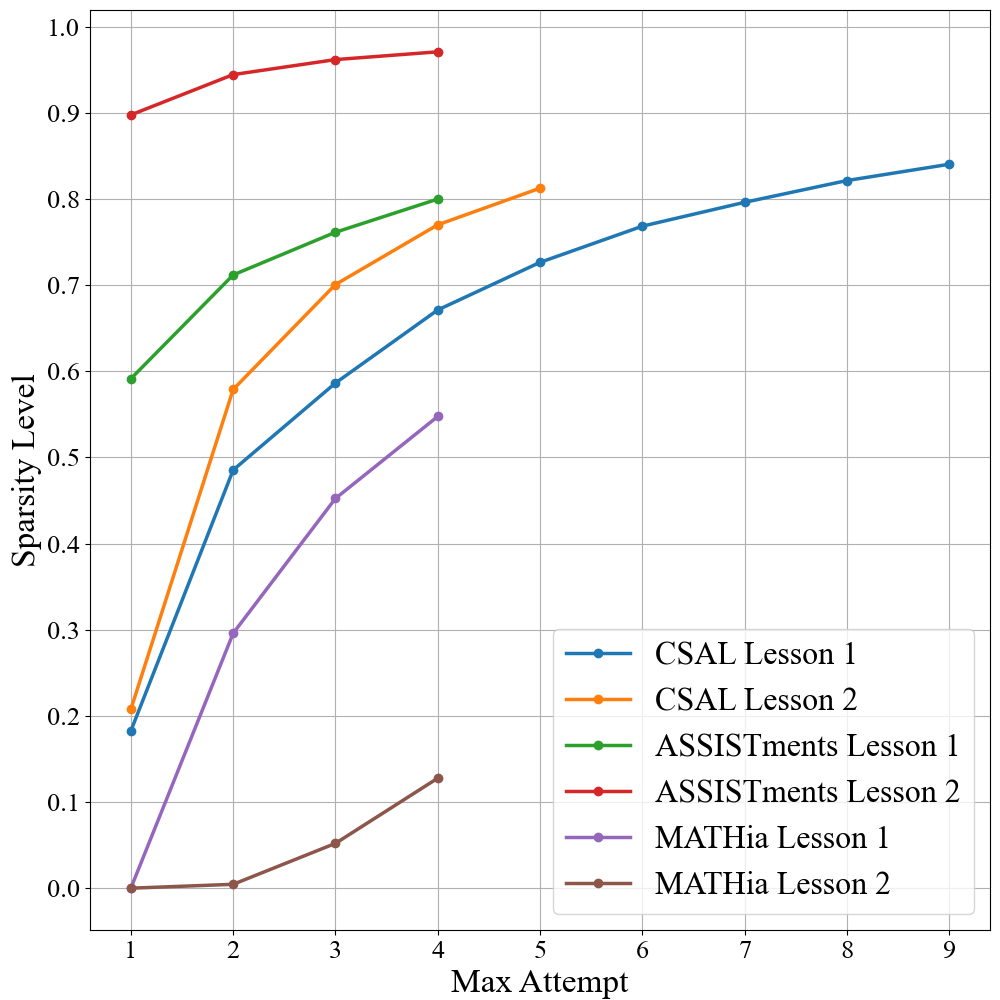}
\captionof{figure}{Data sparsity levels} \label{fig:sparsity}
\end{minipage} 
\vspace{0.5pt}
\begin{adjustwidth}{0pt}{0pt}
-er initial sparsity level, with the former experiencing a sharp increase and the latter following a more gradual trajectory as attempts progress. Particularly, ``CSAL Lesson 1'' records the maximum number of attempts observed for this class. The distinct sparsity patterns observed in Fig.~\ref{fig:sparsity} highlight the heterogeneity of data completeness and the extent of missingness across different lesson datasets. 
\end{adjustwidth}
\begin{figure}
\centering
\includegraphics[width=1\textwidth]{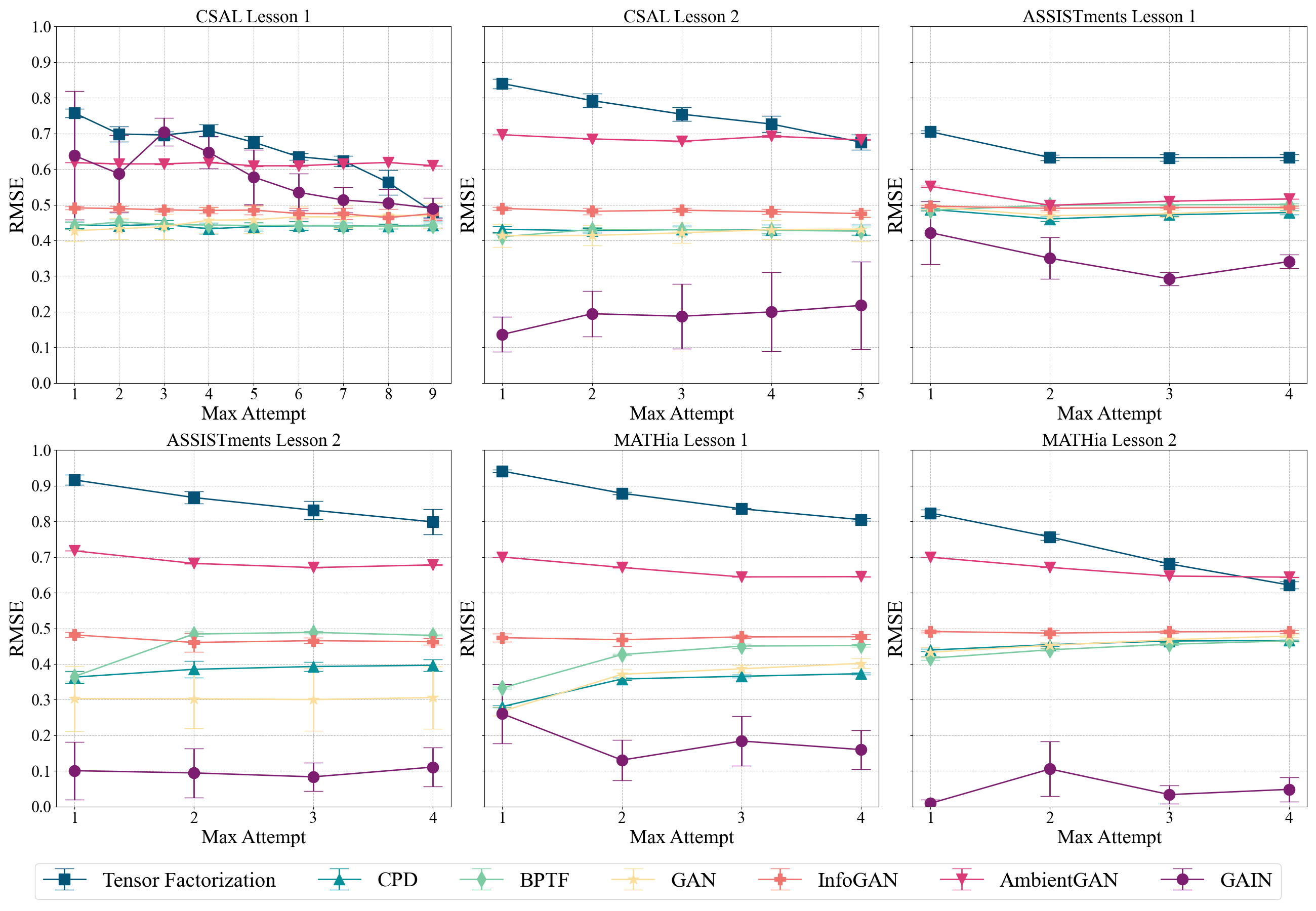}
\caption{RMSE performance in different models for all lessons dataset} \label{fig:performance}
\end{figure} 

\textbf{Models' Imputation Performance.} \textbf{RQ1} investigates how effectively the GAIN-based method imputes sparse learning performance data in ITSs compared to established baseline methods. This question is addressed by the following results. Fig.~\ref{fig:performance} presents the RMSE results of imputations performed by various models on sparse learning performance data across lessons. GAIN significantly outperforms other baseline models across most datasets, particularly in the context of ASSISTments and MATHia lessons, as well as in certain CSAL lessons, save for an exception in CSAL Lesson 1. In CSAL Lesson 1, GAIN's RMSE is lower than that of BPTF, GAN, and InfoGAN, and its extended error bars indicate a comparative decrease in imputation precision and consistency. Remarkably, by the 9th attempt, GAIN's RMSE approaches the minimal value, signifying high accuracy in data imputation toward the higher max attempt. The unique case of CSAL Lesson 1 underscores complex data or model interactions that merit in-depth research to unravel the specific factors influencing its imputation challenges. Additionally, the GAN model exhibits performance surpassing that of other models, albeit slightly less robust than GAIN, while CPD and BPTF demonstrate competitive capabilities. 

\textbf{Stability of Imputation with Varying Attempts.} \textbf{RQ2} examines how the stability of the GAIN-based model’s performance varies with changes in the number of attempts influencing the sparsity levels of learning performance data, as demonstrated by the subsequent results. Despite its overall superior performance, GAIN exhibits greater variance in its results, as indicated by the longer error bars (see Fig. \ref{fig:performance}), which suggests less stability in its data imputation. The heightened variance suggests that while GAIN generally delivers superior imputation, its consistency is compromised under certain data conditions, possibly requiring additional tuning or pre-processing to stabilize its performance. Moreover, the comparative analysis of other baseline models like Tensor Factorization and CPD indicates a lower variance, suggesting that these may provide more reliable imputations in certain contexts, despite not always achieving the lowest RMSE. 

\noindent % Ensures no indentation
\begin{minipage}[t]{0.5\textwidth}
\vspace{0pt} % Aligns the top of the minipage with the top of the right side
\textbf{Iterative Changes of RMSE for GAIN.} As depicted in Fig.~\ref{fig:iteration}, the RMSE trajectory during the example training stage demonstrates the evolution of GAIN's imputation performance across various lessons with each iteration. Implementing an early stopping criterion is essential when the model exhibits satisfactory performance, typically when the RMSE approximates 0.1, to prevent overfitting. Initially, there is a pronounced reduction in RMSE across all lessons, signaling GAIN's rapid enhancement in accuracy. Particularly, ``MATHia Lesson 1'', ``MATHia Lesson 2'', and ``ASSISTments Lesson 1'' exhibit the most considerable decrease, achiev- 
\end{minipage}%
\hspace{0.1cm} % Adds horizontal space between the minipages
\begin{minipage}[t]{0.5\textwidth}
\vspace{2pt} % Aligns the top
\includegraphics[width=1\textwidth]{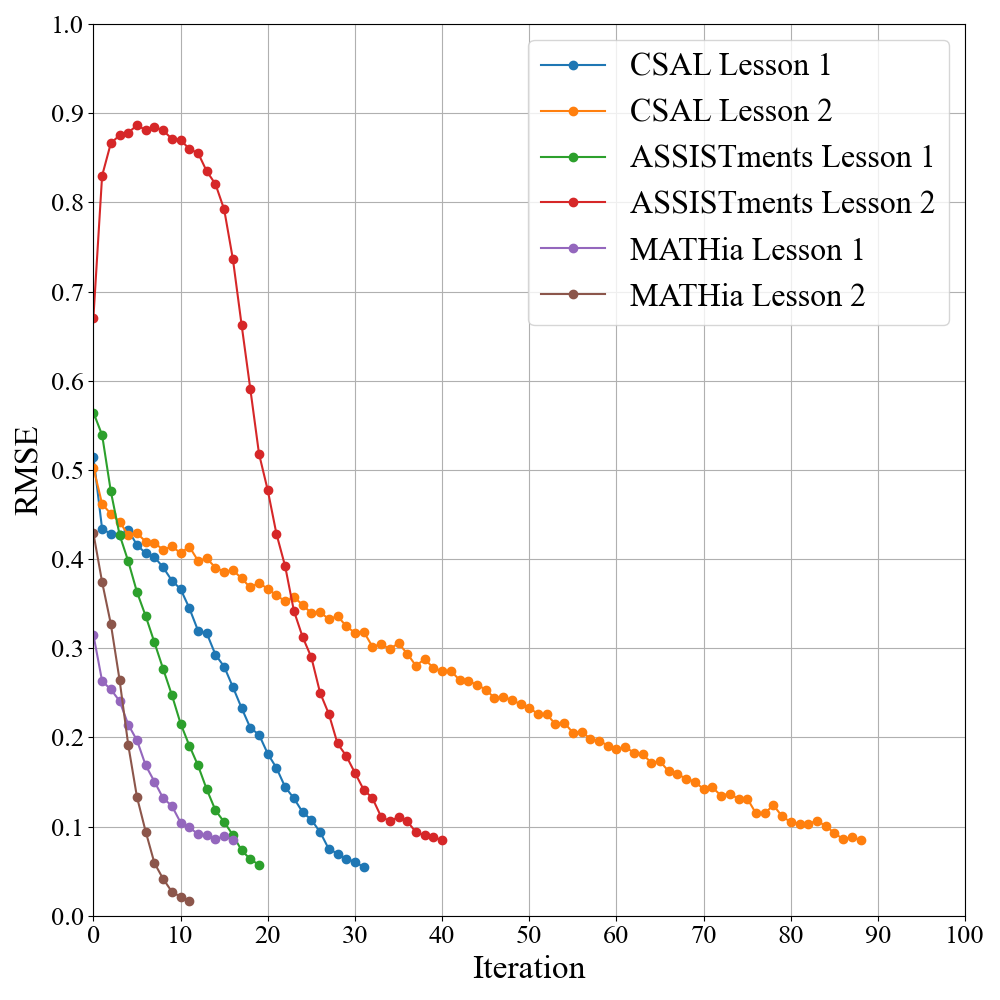}
\captionof{figure}{Data sparsity levels} \label{fig:iteration}
\end{minipage} 
\vspace{0.5pt}
\begin{adjustwidth}{0pt}{0pt}
-ing optimal RMSE levels within fewer than 20 iterations. The prolonged convergence for ``CSAL Lesson 2'' beyond 40 iterations implies that additional gains in accuracy are marginal, prompting considerations for early stopping to optimize computational resources. The variability in the number of iterations required to reach convergence further underscores the diversity of the underlying data distributions. These findings illuminate the complexity of learning performance patterns and the distinctive characteristics of knowledge acquisition that GAIN captures, albeit with varying rates of convergence.
\end{adjustwidth}

\begin{figure}
\includegraphics[width=\textwidth] {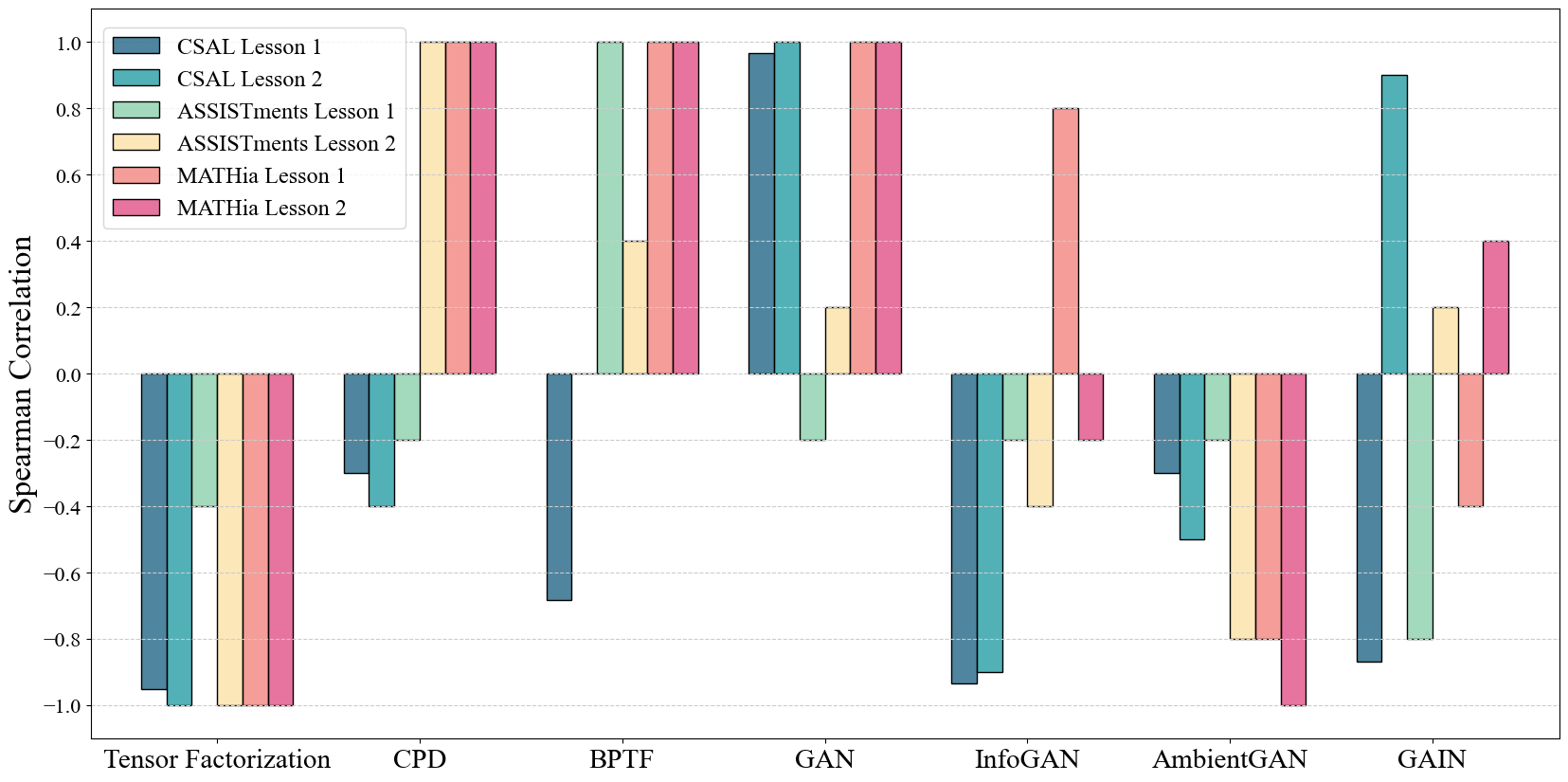}
\caption{Spearman correlation of RMSE with varying attempts.} \label{fig:spearman}
\end{figure} 

\textbf{Spearman Correlation.} Fig.~\ref{fig:spearman} provides a comparison of the Spearman correlation coefficients, quantifying the relationship between RMSE values and the varying maximum number of attempts across various models applied to different lessons. A positive correlation signifies that the model’s RMSE rises as the maximum number of attempts increases, which aligns with a trend of rising sparsity levels, thereby indicating a decline in model performance. Conversely, negative values suggest that the RMSE does not necessarily rise with sparsity, potentially indicating a model's higher performance to missing data. For tensor factorization methods, the prevalent negative correlations across the lessons suggest that their RMSE tends to decrease alongside sparsity. CPD exhibits varied outcomes, with certain lessons reflecting slight positive correlations, while others show negative, indicating inconsistent behavior across different datasets. BPTF predominantly shows positive correlations, with the exception of the CSAL lessons, suggesting a general tendency for model performance to decrease with sparsity within these contexts. For the majority of lessons, GAN demonstrate positive correlations, signaling weaker performance as sparsity enlarges. InfoGAN mostly records negative values, suggesting improved performance in the face of increasing sparsity for most lessons. AmbientGAN consistently exhibits negative correlations for all lessons, implying an increase in model performance despite greater sparsity. GAIN's results are varied, reflecting its nuanced response to the distinct traits of each dataset. 

\subsection{Limitations and Future Works} 

However, the exploration into the adaptive mechanisms of GAIN also highlights areas for future research, particularly in refining model architecture and expanding model explainability to better understand the underlying imputation processes. As educational data continues to grow in complexity and size, further advancements in generative models will be crucial in fully harnessing the potential of generative AI to transform AI-based educational systems.

Future research will delve deeper into the analysis of tensors imputed by GAIN and other methods, focusing on the following areas: 
\begin{itemize}
    \item [$\bullet$] Enhanced evaluation of imputed data is imperative, with an emphasis on comparing imputed and original data patterns. 
    \item [$\bullet$] The use of synthetic datasets, with their adjustable sparsity levels and known ground truth, will further facilitate the evaluation process. 
    \item [$\bullet$] The exploration of generative AI for educational data imputation is still evolving. A systematic comparison of various generative models, including Autoencoders (AE), Variational Autoencoders (VAE), and their interpretable counterparts such as adversarial AEs and Denoising Autoencoders (DAE) \cite{vincent2008extracting}, is planned. 
    \item [$\bullet$] A concerted effort will also be made to distinguish between the semantics of zero values and \(NaN\) values, enhancing the quality of the imputation for learning performance data.
    \item [$\bullet$] An ablation study can verify the effectiveness of the new GAIN-based architecture for imputing sparse learning performance data. By iteratively testing variants by removing or replacing components such as convolutional layers, the hint mechanism, or the least squares loss function, and comparing their performance using key metrics like RMSE, we can identify the contributions of individual components and refine the architecture for optimal effectiveness. 
    \item [$\bullet$] Another potential is the application of generative models in data imputation, which can facilitate ITSs in tracing and predicting learning performance data, especially in real-time and dynamic learning environments. The emerging generative language models, with their reasoning capabilities and the powerful computational abilities of deep generative models, can potentially lead to new advancements in AI-educational applications and research \cite{zhang2024predicting,zhang2024spl}. 
\end{itemize}

\section{Conclusion} 

In this study, we propose a GAIN-based method for imputing sparse learning performance data from ITSs. Our systematic comparison with tensor factorization and other GAN-based methods shows that GAIN not only surpasses these traditional models in terms of imputation accuracy but also demonstrates remarkable adaptability across various educational datasets. However, GAIN's performance is marked by increased variance and diminished stability in data imputation, influenced by varying levels of data sparsity and not uniformly consistent across different lessons. Furthermore, the initial tensor-based representation within a 3D tensor space preserves the original sequence effects and structure, which, when combined with GAIN's use of CNN for its input and output layers, effectively bridges the gap between employing generative AI models for imputing sparse learning performance data and retaining essential temporal educational dynamics. The success of GAIN in this context lays the groundwork for more robust educational data analytics, enhancing decision-making in educational settings, especially in ITSs. This study significantly enriches the application of GAIN in the fields of learning engineering and learning science. 

\subsubsection{Acknowledgements} This paper is grateful to Prof. Philip I. Pavlik Jr. from the University of Memphis and Prof. Shaghayegh Sahebi from the University at Albany - SUNY for their invaluable assistance with tensor factorization in the early stages of this research. Additionally, we extend our thanks to Prof. Arthur C. Graesser, also from the University of Memphis, for his insightful communications that significantly enriched our understanding and inspired deeper analytical thinking. 


\printbibliography

\end{document}